\documentclass[journal]{IEEEtran}  

\usepackage[shortlabels]{enumitem}
\usepackage{comment}
\usepackage{graphicx,epsfig}
\usepackage{color}

\usepackage{amsbsy,amscd,amsfonts,amsgen,amsmath,amsopn,amssymb,amstext,amsthm,amsxtra}
\usepackage{graphicx}
\usepackage{subfigure}
\usepackage{verbatim}
\usepackage{url}
\usepackage{color}
\usepackage{comment}
\usepackage{tikz}
\usepackage{float}
\usepackage{mathtools}

\DeclarePairedDelimiter\floor{\lfloor}{\rfloor}

\usepackage{amsmath}

\newcommand{\blue}[1]{ {#1}}

\begin{document}

\title{Winning Isn't Everything: Enhancing Game Development with Intelligent Agents}

    \author{Yunqi Zhao,* Igor Borovikov,* Fernando~de~Mesentier~Silva,* Ahmad Beirami,*\\
    Jason Rupert, Caedmon Somers,
    Jesse Harder,
    John Kolen, Jervis Pinto, Reza~Pourabolghasem, \\James Pestrak, Harold Chaput, Mohsen~Sardari, Long Lin, Sundeep Narravula, Navid Aghdaie, and Kazi Zaman\vspace{-.1in}

\thanks{This paper was presented in part at
14th AAAI Conference on Artificial Intelligence and Interactive Digital Entertainment (AIIDE 18) ~\cite{Fernando-A-star}, NeurIPS 2018 Workshop on Reinforcement Learning under Partial Observability~\cite{Igor-NeurIPS18, Yunqi-NeurIPS18}, AAAI 2019 Workshop on Reinforcement Learning in Games (Honolulu, HI)~\cite{AAAI19-RLG}, AAAI 2019 Spring Symposium on Combining Machine Learning with Knowledge Engineering~\cite{AAAI-Make}, ICML 2019 Workshop on Human in the Loop Learning~\cite{ICML-HILL}, ICML 2019 Workshop on Imitation, Intent, and Interaction~\cite{MAL-team-sports}, The 23rd Annual Signal and Image Sciences Workshop at Lawrence Livermore National Laboratory, 2019 \cite{LLNL_CASIS_2019}. The code for some of the experiments is available on GitHub~ \cite{ibor_github_jun2019}.}

\thanks{J. Rupert and C. Somers are with EA Sports, Electronic Arts, 4330 Sanderson Way, Burnaby, BC V5G 4X1, Canada. }
\thanks{The rest of the authors are with EA Digital Platform -- Data \& AI, Electronic Arts, Redwood City, CA 94065 USA.}
\thanks{*Y. Zhao, I. Borovikov, FDM Silva and A. Beirami contributed equally to this paper. e-mails: \{yuzhao, iborovikov, fdemesentiersilva\}@ea.com, ahmad.beirami@gmail.com.}
}

\maketitle

\begin{abstract}  
Recently, there have been several high-profile achievements of agents learning to play games against humans and beat them.  
In this paper, we study the problem of training intelligent agents in service of game development. Unlike the agents built to ``beat the game'', our agents aim to produce human-like behavior to help with game evaluation and balancing. We discuss two fundamental metrics based on which we measure the human-likeness of agents, namely skill and style, which are multi-faceted concepts with practical implications outlined in this paper. We report four case studies in which the style and skill requirements inform the choice of algorithms and metrics used to train agents; ranging from A* search to state-of-the-art deep reinforcement learning. We, further, show that the learning potential of state-of-the-art deep RL models does not seamlessly transfer from the benchmark environments to target ones without heavily tuning their hyperparameters, leading to linear scaling of the engineering efforts and computational cost with the number of target domains.
\end{abstract}


\section{Introduction}
The history of artificial intelligence (AI) can be mapped by its achievements playing and winning various games. From the early days of Chess-playing machines to the most recent accomplishments of Deep Blue~\cite{deep-blue}, AlphaGo~\cite{alpha-go}, and AlphaStar~\cite{AlphaStar}, {game-playing} AI\footnote{We refer to game-playing AI as any artificial intelligence solution that powers an agent in the game to simulate a player (user). This can range from rule-based agents to the state-of-the-art deep reinforcement learning agents.} has advanced from competent, to competitive, to champion in even the most complex games.
Games have been instrumental in advancing AI, and most notably in recent times through Monte Carlo tree search and \blue{deep} reinforcement learning (RL). 

\begin{figure}[t]
  \centering
  \includegraphics[width=0.7\linewidth]{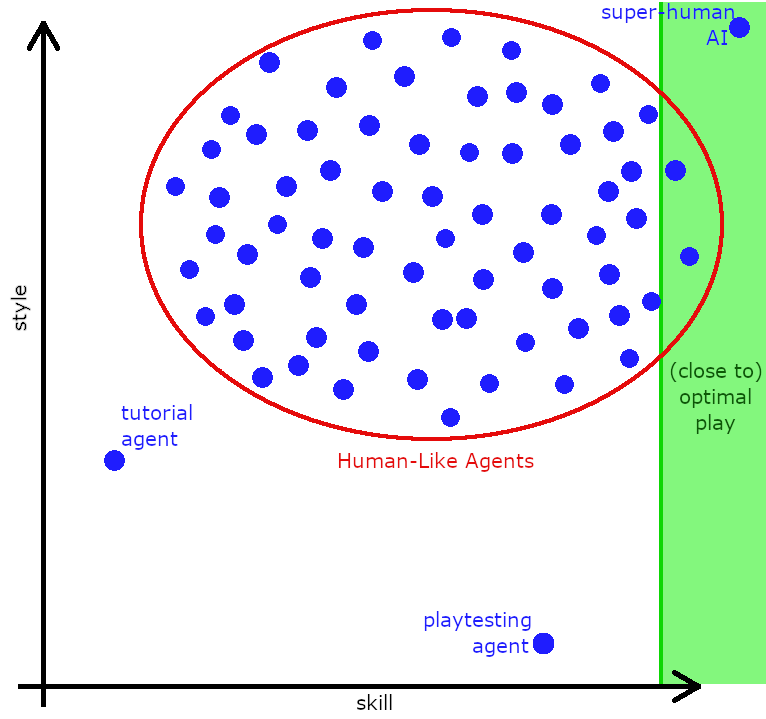}
  \caption{\blue{A depiction of the possible ranges of AI agents and the possible tradeoff/balance between skill and style. In this tradeoff, there is a region that captures human-like skill and style. AI Agents may not necessarily land in the human-like region. High-skill AI agents land in the green region while their style may fall out of the human-like region.} 
  }
  \vspace{-.15in}
  \label{Figure:agent_performance_x_style}
\end{figure}

\blue{Complementary to these great efforts on training high-skill gameplaying agents, at Electronic Arts, our primary goal is to train agents that assist in the game design process, which is iterative and laborious. The complexity of modern games steadily increases, making the corresponding design tasks even more challenging. To support designers in this context, we train game-playing AI agents as user simulators to perform tasks ranging from automated playtesting to interaction with human players tailored to enhance game-play immersion.}

\blue{
To approach the challenge of creating agents that can generate meaningful interaction data to inform game developers, we propose simple techniques to model different user behaviors. Each agent has to strike a different balance between {\em style} and {\em skill}. 
We define skill as how efficient the agent is at completing the task it is designed for.  Style is vaguely defined as how the player engages with the game and what makes the player enjoy their game-play. Defining and gauging skill is usually much easier than that of style. Gameplay style in itself is a complex concept, spawning a field of its own. While a comprehensive study of style is outside the scope of this work, we refer the interested reader to ~\cite{drachen2009playerstyle, gow2012unsupervisedplayerstyle} on modeling player style. In this work, we attempt to evaluate style of an artificial agent using statistical properties of the underlying simulator model.
}

\blue{One of the most crucial tasks in game design is the process of playtesting. 
Game designers usually rely on playtesting sessions and feedback they receive from playtesters to make design choices in the game development process.
Playtesting is performed to guarantee quality game-play that is free of game-breaking exceptions (e.g., bugs and glitches) and delivers the experience intended by the designers. Since games are complex entities with many moving parts, solving this multi-faceted optimization problem is even more challenging. An iterative loop where data is gathered from the game by one or more playtesters, followed by designer analysis is repeated many times throughout the game development process.}

\blue{
To mitigate this expensive process, one of our major efforts is to implement agents that can help automate aspects of playtesting. These agents are meant to play through the game, or a slice of it, trying to explore behaviors that can generate meaningful data to assist in answering questions that designers pose. These can range from exhaustively exploring certain sequences of actions, to trying to play a scenario from start to finish with the shortest sequence of actions possible. We showcase use-cases focused on creating AI agents to playtest games at Electronic Arts and discuss the related challenges.}

\blue{Another key task in game development is the creation of in-game characters that interact with real human players.  Agents must be trained and delicate tuning has to be performed to guarantee quality experience (e.g., engagingness and humanness). An AI adversary with an unreasonably fast reaction time can be deemed unfair rather than challenging. On the other hand, a pushover agent might be an appropriate introductory opponent for novice players, while it fails to retain player interest after a handful of matches. While traditional AI solutions can provide excellent experiences for the players, it is becoming increasingly more difficult to scale those traditional solutions up as the game worlds are becoming larger and the content is becoming dynamic.}

\blue{
In our experience, as Fig. \ref{Figure:agent_performance_x_style} shows, we have observed that there is a range of style/skill pairs that are achievable by human players, and hence called human-like. 
High-skill game-playing agents may have an unrealistic style rating if they rely on high computational power and memory size, and reaction times unachievable by humans. Evaluation of techniques to emulate human-like behavior have been presented~\cite{ortega2013imitating}, but measuring non-objective metrics such as fun and immersion is an open research question~\cite{fun-in-games, immersion-in-games}. Further, we cannot evaluate player engagement prior to the game launch, so we rely on our best approximation: {\em designer feedback}. Through an iterative process, designers evaluate the game-play experience by interacting with the agents to measure whether the intended game-play experience is provided.}

\blue{These challenges each require a unique equilibrium between style and skill. Certain agents could take advantage of superhuman computation to perform exploratory tasks, most likely relying more heavily on skill. Others need to interact with human players, requiring a style that won't break player immersion. Then there are agents that need to play the game with players cooperatively, which makes them rely on a much more delicate balance that is required to pursue a human-like play style. Each of these  individual problems call for different approaches and have significant challenges. Pursuing human-like style and skill can be as challenging (if not more) than achieving high performance agents.}

Finally, training an agent to a specific need is often more efficient than achieving such solution through high-skill AI agents. This is the case, for example, when using game-playing AI to run multiple playthroughs of a specific in-game scenario to trace the origin of an unintended game-play behavior. In this scenario, an agent that would explore the game space would potentially be a better option than one that reaches the goal state of the level more quickly. Another advantage in creating specialized solutions (as opposed to artificial general intelligence) is the cost of implementation and training. The agents needed for these tasks are, commonly, of less complexity in terms of needed training resources.

{To summarize,} we mainly pursue two use-cases for having AI agents enhance the game development process.
\begin{enumerate}
    \item {\em playtesting AI agents} to provide feedback during the game's development.
    
    \item \blue{{\em game-playing AI agents} to interact with real human players to shape their game-play experience.}
\end{enumerate}

The rest of the paper is organized as follows. In Section~\ref{sec:related_work}, we review the related work on training agents for playtesting and NPCs. In Section~\ref{sec:pipeline}, we describe our training pipeline.
\blue{In Sections~\ref{sec:feedback} and~\ref{sec:NPC}, we provide four case studies that cover playtesting and game-playing, respectively. These studies are performed to help with the development process of multiple games  at Electronic Arts. These games vary considerably in many aspects, such as the game-play platform, the target audience, and the engagement duration. The solutions in these case studies were created in constant collaboration with the game designers. The first case study in Section~\ref{sec:sims-mobile}, which covers game balancing and playtesting was done in conjunction with the development of The Sims Mobile. The other case studies are performed on games that are still under development at the time this paper was written. Hence, we had to omit specific details purposely to comply with company confidentiality.}
 Finally, the concluding remarks are provided in Section~\ref{sec:conclusion}.

\section{Related Work}
\label{sec:related_work}

\subsection{Playtesting AI agents (user simulators)}

To validate their design, game designers conduct playtesting sessions. Playtesting consists of having a group of players interact with the game in the development cycle to not only gauge the engagement of players, but to discover states that result in undesirable outcomes. As a game goes through the various stages of development, it is essential to continuously iterate and improve the relevant aspects of the game-play and its balance. Relying exclusively on playtesting conducted by humans can be costly and inefficient. Artificial agents could perform much faster play sessions, allowing the exploration of much more of the game space in much shorter time. 

When applied to playing computer games, RL assumes that the goal of a trained agent is to achieve the best possible performance with respect to clearly defined rewards while the game itself remains fixed for the foreseen future. In contrast, during game development the objectives and the settings are quite different and vary over time. Agents can play a variety of roles with rewards that are not obvious to define formally, e.g., an objective exploring a game level is different from defeating all adversaries. In addition, the environment often changes between the game builds. In such settings, it is desirable to quickly train agents that help with automated testing and data generation for the game balance and feature evaluation. It is also desirable that the agent be re-usable despite new aesthetics and game-play features.
A strategy of relying on increasing computational resources combined with substantial engineering efforts to train agents in such conditions is far from practical and calls for a different approach.

The idea of using artificial agents for playtesting is not new.
Algorithmic approaches have been proposed to address the issue of game balance, in board games~\cite{de2017contemporaryboardgameai,hom2007automatic} and card games~\cite{r2014,mahlmann2012evolving, silva2019evolving}. 
More recently, Holmgard {\em et al.}~\cite{holmgard}, as well as, Mugrai {\em et al.}~\cite{mugrai2019automated} built variants of MCTS to create a player model for AI Agent based playtesting. \blue{Guerrero-Romero {\em et al.} created different goals for general game-playing agents in order to playtest games emulating players of different profiles ~\cite{guerrero2018using}}. These techniques are relevant to creating rewarding mechanisms for mimicking player behavior.
AI and machine learning can also play the role of a co-designer, making suggestions during development process~\cite{yannakakis2014mixed}. 
Tools for creating game maps~\cite{liapis2013sentient} and level design~\cite{smith2010tanagra,shaker2013ropossum} are also proposed. 
See~\cite{search-based-generation,pcgml} for a survey of these techniques in game design.

In this paper, we describe our framework that supports game designers with automated playtesting. This also entails a training pipeline that universally applies this framework to a variety of games.
We then provide two case studies that entail different solution techniques.

\subsection{Game-playing AI agents (Non-Player Characters)}

Game-playing AI has been a main constituent of games since the dawn of video gaming. \blue{Analogously, games, given their challenging nature, have been a target for AI research~\cite{yannakakis2018artificial}. Over the years, AI agents have become more sophisticated and have been providing excellent experiences to millions of players as games have grown in complexity.}
Scaling traditional AI solutions in ever growing worlds with thousands of agents and dynamic content is a challenging problem calling for alternative approaches.

\blue{The idea of using machine learning for game-playing AI dates back to}
 Arthur Samuel~\cite{samuel-checkers}, who applied some form of tree search combined with basic reinforcement learning to the game of checkers. His success motivated researchers to target other games using machine learning, and particularly reinforcement learning.

IBM Deep Blue followed the tree search path and was the first artificial game agent who beat the chess world champion, Gary Kasparov~\cite{deep-blue}.
A decade later, Monte Carlo Tree Search (MCTS)~\cite{MCTS,UCT} was a big leap in AI to train game agents.
MCTS agents for playing Settlers of Catan were reported in~\cite{settlersszita2009monte,settlerschaslot2008monte} and shown to beat previous heuristics. Other work compares multiple approaches of agents to one another in the game Carcassonne on the two-player variant of the game and discusses variations of MCTS and Minimax search for playing the game~\cite{Carcassonneheyden2009implementing}. MCTS has also been applied to the game  of 7 Wonders~\cite{7wonders} and Ticket to Ride~\cite{MCTSTicketToRide}. \blue{Furthermore, Baier {\em et al.} biased MCTS with a player model, extracted from game-play data, to have an agent that was competitive while approximating human-like play~\cite{baier2018emulating}.} 
Tesauro~\cite{TDGammon}, on the other hand, used TD-Lambda to train Backgammon agents at a superhuman level. The impressive recent progress on RL to solve video games is partly due to the advancements in processing power and AI computing technology.\footnote{The amount of AI compute has been doubling every 3-4 months in the past few years~\cite{openai-compute}.} More recently, \blue{following the success stories in deep learning, deep Q networks (DQNs) use deep neural networks as function approximators within Q-learning~\cite{DQN}. DQNs can use convolutional function approximators as a general representation learning framework from the pixels in a frame buffer without need for task-specific feature engineering.} 

DeepMind remarried the approaches by combining DQN with MCTS to create AI agents that play Go at a superhuman level~\cite{alpha-go}, and solely via self-play~\cite{alpha-go-zero,alpha-zero}.
Subsequently, OpenAI researchers showed that a policy optimization approach with function approximation, called Proximal Policy Optimization (PPO)~\cite{PPO}, would lead to training agents at a superhuman level in Dota 2~\cite{openai-dota2}. \blue{Cuccu {\em et al.} proposed learning policies and state representations individually, but at the same time, and did so using two novel algorithms~\cite{cuccu2019playing}.  With such approach they were able to play Atari games with neural networks of 18 neurons or less.} Recently, progress was reported by DeepMind on StarCraft II, where AlphaStar was unveiled to play the game at a high-competitive human level by combining several techniques, including attention networks~\cite{AlphaStar}.

\begin{figure}
  \centering
  \includegraphics[width=\linewidth]{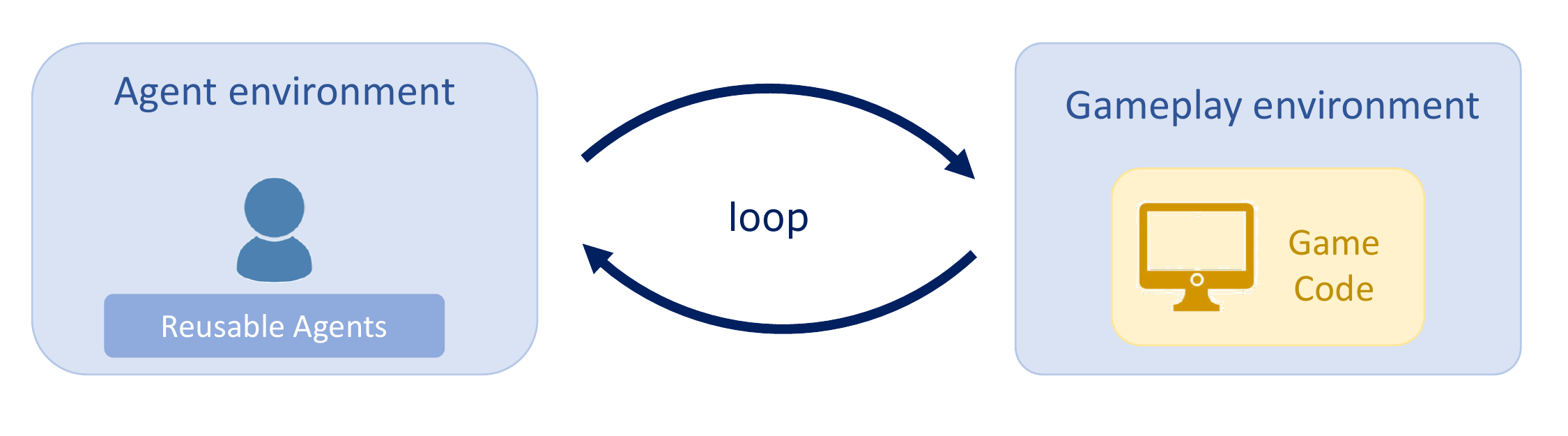}
  \caption{The AI agent training pipeline, \blue{which consists of two main components, game-play environment and agent environment. Agents submit actions to the game-play environment and receive the next state.}}
  \label{Figure:training_pipeline}
\end{figure}

\section{Training Pipeline}
\label{sec:pipeline}
\blue{To train AI agents efficiently, we developed a unified training pipeline applicable to all of EA games, regardless of platform and genre. In this section, we present our training pipeline that is used for solving the case studies presented in the sections that follows.}

\subsection{Gameplay and Agent Environments}
The AI agent training pipeline, which is depicted in Fig.~\ref{Figure:training_pipeline}, consists of two key components:
\begin{itemize}
\item {\it Gameplay environment} refers to the simulated game world that executes the game logic with actions submitted by the agent every timestep and produces the next state.
\item {\it Agent environment} refers to where the agent interacts with the game world. The agent observes the game state and produces an action. It is where training occurs.
\end{itemize}
In practice, the game architecture can be complex and infeasible for the game to directly communicate the complete state space at every timestep.
To train artificial agents, we create an interface between game-play environment
and learning environment.\footnote{These environments may be physically separated, and hence, we prefer a client that supports fast cloud execution, and is not tied to frame rendering.} The interface extends OpenAI Gym~\cite{openai-gym} and supports actions that take arguments, which is necessary to encode action functions and is consistent with PySC2~\cite{starcraft2}. 
We also adapt Dopamine~\cite{dopamine} to this pipeline to make DQN~\cite{DQN}, Rainbow~\cite{rainbow} and PPO~\cite{schulman2017proximal} agents available for training in the game. 
Additionally, we add support for more complex preprocessing of game state features other than the usual frame buffer stacking.

\subsection{State Abstraction}
The use of frame buffer as an observation of the game state has proved advantageous in eliminating the need for manual feature engineering in Atari games~\cite{DQN}. However, to enable efficient training (using imitation learning or reinforcement learning) in a fast-paced game development process, the drawbacks of using frame buffer outweigh its advantages. The main considerations which we take into account when deciding in favor of a lower-dimensional engineered representation of game state are:
\begin{enumerate}[(a)]
\item During almost all stages of development, the game parameters are evolving. In particular, the art may change at any moment and the look of already learned environments can change overnight. It is desirable to train agents using more stable features that can transfer to new environments with little need for retraining.

\item State abstraction allows us to train much smaller models (networks) because of the smaller input size and use of carefully engineered features. This is critical for real time application environments where rendering, animation and physics are occupying much of the GPU and CPU power at inference time.

\item
In playtesting, game-play environment and learning environment may reside in physically separate nodes. Naturally, RL state-action-reward loop in such requires a lot of network communication. Frame buffers would significantly increase this communication cost whereas derived game state features enable more compact encoding.

\item  Obtaining an artificial agent in a reasonable time usually requires that the game be clocked at a rate much higher than the usual game-play speed. As rendering each frame takes significant time, overclocking with rendering enabled is not practical. Additionally, moving large amounts of data from GPU to main memory drastically slows down execution and can potentially introduce simulation artifacts, by interfering with the target timestep rate.

\item Last but not least, we can leverage having access to the game code to have the game engine distill a compact state representation and pass it to the agent environment. By doing so we also have better hope of learning in environments where the pixel frames only contain partial information about the the state space.
\end{enumerate}

Feature selection for the compact state representation may require some engineering efforts, but it is straightforward after familiarization with game mechanics. It is often similar to that of traditional game-playing AI, which is informed by the game designer. We remind the reader that our goal is not to train agents to win, but to simulate human-like behavior, so we train on information accessible to human players.

In the rest of this paper, we present four case studies on training intelligent agents for game development; two of which focusing on playtesting AI agents, and the next two on gameplaying AI agents.

\section{\blue{Playtesting AI Agents}}
\label{sec:feedback}


\subsection{\blue{Measuring player experience for different player styles}}
\label{sec:sims-mobile}

In this section, we consider the early development of The Sims Mobile, whose game-play is about ``emulating life'': players create avatars, called Sims, and conduct them through a variety of everyday activities. In this game, there is no single predetermined goal to achieve. Instead, players craft their own experiences, and the designer's objective is to evaluate different aspects of that experience.
In particular, each player can pursue different careers, and each will have a different experience and trajectory in the game. 
\blue{In this specific case study, the designer's goal is to evaluate if the current tuning of the game achieves the intended balanced game-play experience across different careers. For example, different careers should prove similarly difficult to complete. }
We refer the interested reader to \cite{Fernando-A-star} for a more comprehensive study of this problem.

The game is \blue{single-player, deterministic, real-time,} fully observable and the dynamics are fully known. \blue{We also have access to the complete game state, which is composed mostly of character and on-going action attributes.} This simplified case allows for the extraction of a lightweight model of the game \blue{(i.e., state transition probabilities).} While this requires some additional development effort, we can achieve a dramatic speedup in training agents by avoiding (reinforcement) learning and resorting to planning techniques instead.

In particular, we use the A* algorithm for \blue{the simplicity of proposing a heuristic that can be tailored to the specific designer need} by exploring the state transition graph. The customizable heuristics and the target states corresponding to different game-play objectives, \blue{which represent the style we are trying to achieve}, provide sufficient control to conduct various experiments and explore multiple aspects of the game. 

\blue{Our heuristic for the A* algorithm is the weighted sum of the 3 main parameters that contribute for career progression: career level, current career experience points and amount of completed career events. These parameters are directly related. To gain career levels players have to accumulate career experience points and to obtain experience, players have to complete career events. The weights are attributed based on the order of magnitude each parameter has. Since levels are the most important, it receives the highest weight. The total completed career events has the lowest weight because it is already partially factored into the career points received so far.}

\begin{figure}
  \centering
  \includegraphics[width=1.0\linewidth]{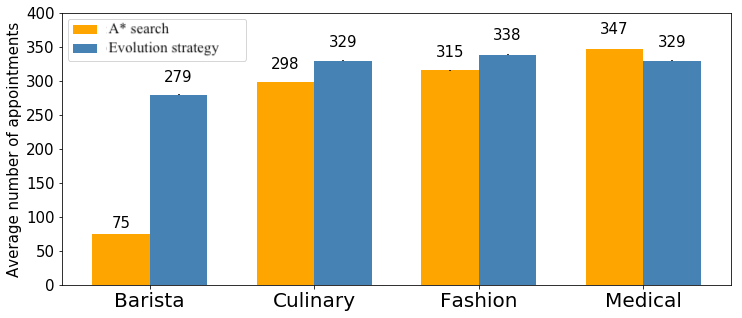}
  \caption{Comparison of the average amount of actions (appointments) taken to complete the career using A* search and evolution strategy adapted from~\cite{Fernando-A-star}.}
  \label{Figure:career_approach_comp}
\end{figure}

\blue{
We also compare A* results to the results from an optimization problem over a subspace of utility-based policies approximately solved with an evolution strategy (ES) \cite{openai-es}. Our goal, in this case, is to achieve a high environment reward against selected objective, e.g., reach the end of a career track while maximizing earned career event points. We design ES objective accordingly. The agent performs an action $a$ based on a probabilistic policy by taking a softmax on the utility $U(a, s)$ measure of the actions in a game state $s$. Utility here serves as an action selection mechanism to compactly represent a policy. In a sense, it is a proxy to a state-action value $Q$-function $Q(a, s)$ in RL. However, we do not attempt to derive utility from Bellman's equation and the actual environment reward $R$. Instead, we learn parameters that define $U(a, s)$ to optimize the environment rewards using the black-box ES optimization technique. In that sense optimizing $R$ by learning parameters of $U$ is similar to Proximal Policy Optimization (PPO), however, in much more constrained settings. To this end, we design utility of an action as a weighted sum of the immediate action rewards $r(a)$ and costs $c(a)$. These are vector-valued quantities and are explicitly present in the game tuning describing the outcome of executing such actions. The parameters evolving by the ES are the linear weights for the utility function $U$ explained below and the temperature of the softmax function. An additional advantage of the proposed linear design of the utility function is a certain level of interpretability of the weights corresponding to the perceived by the agent utilities of the individual components of the resources or the immediate rewards. Such interpretability can guide changes to the tuning data.}

{
Concretely, given the game state $s$, we design the utility $U$ of an action $a$ as
$U(s,a) = r(a)v(s) + c(a)w(s).$
The immediate reward $r(a)$ here is a vector that can include quantities like the amount of experience, amount of career points earned for the action and the events triggered by it. The costs $c(a)$ is a vector defined similarly. The action costs specify the resources required to execute such an action, e.g., how much time, energy, etc. a player needs to spend to successfully trigger and complete the action. The design of the tuning data makes the both quantities $r$ and $c$ only depend on the action itself. Since both - the immediate reward $r(a)$ and $c(a)$ are vector values, the products in the definition of $U$ above are dot products. The vectors $v(s)$ and $w(s)$ introduce dependence of the utility on the current game state and are the weights defining relative contribution of the immediate resource costs and immediate rewards towards the current goals of the agent.

Inferred utilities of the actions depend on the state. Some actions in certain states are more beneficial than in others, e.g., triggering a career event when not having enough resources to complete it successfully. 
The relevant state components $s=(s_1,..,s_k)$ include available commodities like energy and hunger and a categorical event indicator (0 if outside of the event and 1 otherwise) wrapped into a vector. The total number of the relevant dimensions here is $k$. We design the weights $v_a(s)$ and $w_a(s)$ as bi-linear functions with the coefficients ${\bf p}=(p_1,...,p_k)$ and ${\bf q}=(q_1,...,q_k)$ that we are learning: $v_a(s)=\sum_{i=1,..,k} p_i s_i$ and $w_a(s)=\sum_{i=1,..,k} q_i s_i$.

To define the optimization objective $J$, we construct it as a function of the number of successfully completed events $N$ and the number of attempted events $M$. We aim to maximize the ratio of successful to attempted events times total number of successful events in the episode as follows: 
$J(N, M) = N (N + \epsilon) / (M + \epsilon)$,
where $\epsilon$ is a small positive real number to ensure stability when the policy fails to attempt any events. The overall optimization problem is therefore 
$\max_{{\bf p}, \bf{q}} J(N, M)$
subject to the policy parameterized with the parameters $\bf p$ and $\bf q$.

The utility-based ES, as we describe it here, captures the design intention of driving career progression in the game-play by successful completion of career events. Due to the emphasis on the events completion, our evolution strategy setup is not necessarily resulting in an optimization problem equivalent to the one tackled with $A^*$. However, as we discuss below, it has similar optimum most of the time, supporting the design view on the progression. A similar approach works for evaluating relationship progression, which is another important element of the game-play. 
}

We compare the number of actions that it takes to reach the goal for each career in  
Fig.~\ref{Figure:career_approach_comp} as computed by the two approaches. 
\blue{We can see that the more expensive optimization based evolution strategy performs similarly to the much simpler A* search.}

The largest discrepancy arises for the Barista career. This can be from the fact that this career has an action that does not reward experience by itself, but rather enables another action that does it. This action can be repeated often and can explain the high numbers. Also, we observe that in the case of the medical career, the 2,000 node A* cutoff was potentially responsible for the under performance in that solution.

When running the two approaches, we can compare how many sample runs are required to obtain statistically significant results. The A* agent learns a deterministic playstyle, with no variance. We performed 2,000 runs for the evolution strategy, and the agent has a high variance and requires a sufficiently high number of runs to approach a final reasonable strategy~\cite{Fernando-A-star}. 

\blue{In this use case, we were able to use a planning algorithm, A*, to explore the game space to gather data for the game designers to evaluate the current tuning of the game. This was possible due to the goal being straightforward, to evaluate progression in the different careers. With such, the requirements of skill and style for the agent were achievable and simple to model. Over the next use cases, we analyze scenarios that call for different approaches as consequence of having more complex requirements and subjective agent goals.}

\begin{figure*}
    \centering
    \includegraphics[width=0.48\linewidth]{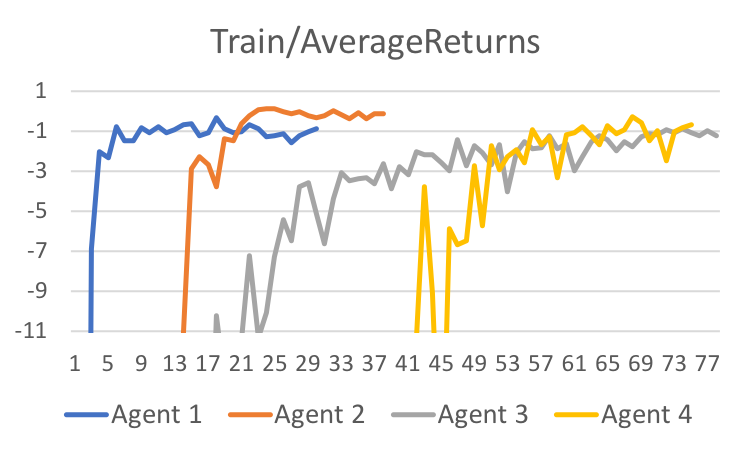}
    \includegraphics[width=0.48\linewidth]{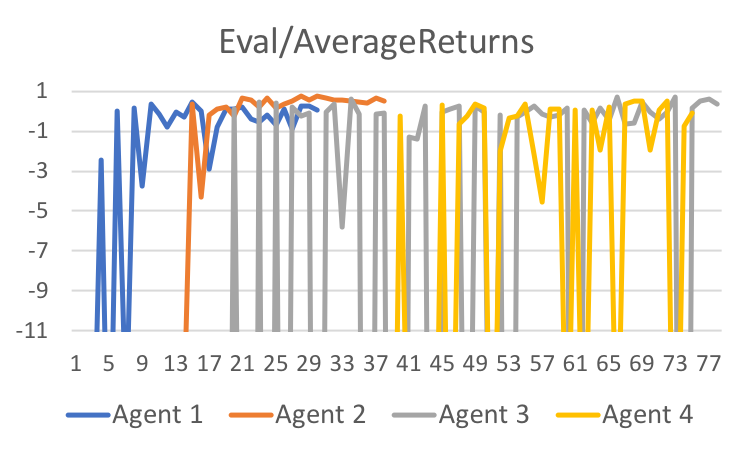}
    \caption{\blue{This plot belongs to Section~\ref{sec:player-progression}.} Average cumulative reward (return) in training and evaluation for the agents as a function of the number of iterations. Each iteration is worth $\sim$60 minutes of game-play. The trained agents are: (1) a DQN agent with complete state space, (2) a Rainbow agent with complete state space, (3) a DQN agent with augmented observation space, and (4) a Rainbow agent with augmented observation space. Augmented space is the space observable by humans in addition to inferred information, which is much smaller than the complete space. }\label{fig:result} 
  \end{figure*}

\subsection{Measuring competent player progression}
\label{sec:player-progression}

\blue{In the next case study, we consider a real-time multi-player mobile game, with a stochastic environment and sequential actions. The game dynamics are governed by a complex physics engine, which makes it impractical to apply planning methods.
This game is more complex than The Sims Mobile in the sense that strategic decision making is required for progression.
}
When the game dynamics are unknown \blue{or  complex}, most recent success stories are based on \blue{model-free} RL (and particularly variants of DQN and PPO). In this section, we show how such model-free control techniques fit into the paradigm of playtesting modern games. 

\blue{
In this game, the goal is to level up and reach milestones in the game. To this end, the players need to make decisions in terms of resource mining and management for different tasks. In the process, the agent needs to perform upgrades that  require certain resources. If such resources are insufficient, a human player will be able to visually discern the validity of such action by clicking on the particular upgrade. 
The designer's primary concern in this case study is to measure how a competent player would progress in the early stages of this game. In particular, players are required to balance resources and make strategic choices that the agent needs to discern as well.
}

\blue{We consider a simplified state space that contains information about the early game, ignoring the full state space.}
The \blue{relevant part of the} state \blue{space} consists of $\sim$50 continuous and $\sim$100 discrete state variables.
The set of possible actions $\alpha$ is a subset of a space $A$, which consists of $\sim$25 action classes, some of which are from a continuous range of possible action values, and some are from a discrete set of action choices. The agent has the ability to generate actions $\alpha \in A$ but not all of them are valid at every game state since $\alpha=\alpha(s, t)$, i.e., $\alpha$ depends on the timestep and the game state. Moreover, the subset $\alpha(s, t)$ of valid actions \blue{may} only partially \blue{be} known to the agent. 
\blue{If the agent attempts to take an unavailable action, such as a building upgrade without sufficient resources, the action will be deemed invalid and no {\em actual} action will be taken by the game server. 
}

While the problem of a huge state space~\cite{poupart-pomdp-continuous,pomdp-continuous,poupart-pomdp-continuous2}, a continuous action space~\cite{silver-cont-action}, and a parametric action space~\cite{stone-parameterized} could be dealt with,
these techniques are not directly applicable to our problem. This is because, as we shall see, some actions will be invalid at times and inferring that information may not be fully possible from the observation space.
Finally, the game is designed to last tens of millions of timesteps, taking the problem of training a functional agent in such an environment outside of the domain of previously explored problems.

We study game progression while taking only valid actions. As we already mentioned, the set of valid actions $\alpha$ \blue{may} not \blue{be} fully determined by the current observation, and hence, we deal with a partially observable Markov decision process (POMDP).
Given the practical constraints outlined above, it is infeasible to apply deep reinforcement learning to train agents in the game in its entirety. 
In this section, we show progress toward training an artificial agent that takes valid actions and progresses in the game like \blue{a competent} human player. \blue{To this end, we wrap this game in the game environment and connect it} to our training pipeline with DQN and Rainbow agents.
\blue{In the agent environment,} we use a \blue{feedforward neural} network with two fully connected hidden layers, \blue{each with 256 neurons followed by} ReLU activation.

\blue{As a first step in measuring game progression,} we \blue{define} an episode by setting an early goal state in the game that takes an expert human player $\sim$5 minutes to reach.
 We let the agent submit actions to the game server every second. \blue{We may have to revisit this assumption for longer episodes where the human player is expected to interact with the game more periodically.}
\blue{We use a simple rewarding mechanism, where} we reward the agent with `+1' when they reach the goal state, `-1' when they submit an invalid action, `0' when they take a valid action, and `-0.1' when they choose the ``do nothing'' action. The game is such that at times the agent has no other valid action to choose, and hence they should choose ``do nothing'', but such periods do not last more than a few seconds \blue{in the early stages of the game, which is the focus of this case study.}

We consider two different versions of the observation space, both extracted from the game engine \blue{(state abstraction)}. The first is what we call the ``naive'' state space. The complete state space contains information that is not straightforward to infer from the real observation in the game and is only used as a baseline for the agent. \blue{In particular, the complete state space also includes the list of available actions at each state.}  The polar opposite of this state space could be called the ``naive'' state space, which only contains straightforward information \blue{that is always shown on the screen of the player.}
The second state space we consider is what we call the ``augmented'' observation space, which contains information from the naive state space and information the agent would infer and retain from current and previous gameplays. \blue{For example, this includes the amount of resources needed for an upgrade after the agent has checked a particular building for an upgrade.
The augmented observation space does not include the set of all available actions, and hence, 
we rely on the game server to validate whether a submitted action is available because it is not possible to encode and pass the set $\alpha$ of available actions. Hence, if an invalid action is chosen by the agent, the game server will ignore the action and will flag the action so that we can provide a `-1' reward.}

We trained four types of agents as shown in Fig.~\ref{fig:result}, where we are plotting the average undiscounted return per episode. By design, this quantity is upper bounded by `+1', which is achieved by taking valid actions until reaching the final goal state. In reality, this may not always be achievable as there are periods of time where no action is available and the agent has to choose the ``do nothing'' action and be rewarded with `-0.1'. Hence, the best a \blue{competent} human player would achieve on these episodes would be around zero.

We see that after a few iterations, both the Rainbow and DQN agents converge to their asymptotic performance values.
The Rainbow agent converges to a better performance level compared to the DQN agent. 
However, in the the transient behavior we observe that the DQN agent achieves the asymptotic behavior faster than the Rainbow agent. \blue{We believe this is due to the fact that hyperparameters of prioritized experience replay~\cite{prioritized-experience-replay} and distributional RL~\cite{distributional-RL} were not tuned.\footnote{\blue{This is consistent with the results of Section~\ref{sec:team-sports}, where Rainbow with default hyperparameters does not outperform DQN either.}} We used the default values that worked best on Atari games with frame buffer as state space.
Extra hyperparameter tuning would have been costly in terms of cloud infrastructure and time for this particular problem since the game server does not allow speedup, building the game takes several hours, and training one agent takes several hours as well.}

As expected, Fig.~\ref{fig:result} shows that the augmented observation space makes the training slower and has worse performance on the final strategy. In addition, the agent keeps attempting invalid actions in some cases as the state remains mostly unchanged after each attempt and the policy is (almost) deterministic. These results in accumulating large negative returns in such episodes which account for the dips in the right-hand-side panel in Fig.~\ref{fig:result} at evaluation time.
The observed behavior drew our attention to the question of whether it is too difficult to discern and keep track of the set of valid actions for a human player as well. In fact, after seeking more extensive human feedback the game designers concluded that better visual cues were needed for a human player on information about valid actions at each state so that the human players could progress more smoothly without being blocked by invalid actions. As next steps, we intend to experiment with shaping the reward function for achieving different play styles to be able to better model different player clusters. Comparison between human play styles and agent emulated styles is discussed in \cite{kemmerling2013making}. We  also intend to investigate augmenting the replay buffer with expert demonstrations for faster training and also for generative adversarial imitation learning~\cite{GAIL} once the game is released and human play data is available.

\blue{
We remark that without state abstraction (streamlined access to the game state), the neural network function approximator used for Q-learning would have needed to discern all such information from the pixels in the frame buffer, and hence we would not have been able to get away with such a simple two-layer feedforward function approximator to solve this problem.
However,}
we observe that the training within the  paradigm of \blue{model-free} RL remains costly. Specifically, even using the complete state space, it takes several hours to train \blue{an agent} that achieves a level of performance expected of a \blue{competent} human player on this relatively short episode of $\sim$5 minutes.
This calls for the exploration of complementary approaches to augment the training process. \blue{In particular, we also would like to streamline this process by training reusable agents and capitalizing on existing human data through imitation learning.}

\section{Game-playing AI}
\label{sec:NPC}
We have shown the value of simulated agents in a fully modeled game, and the potential of training agents in a complex game to model player progression \blue{for game balancing.} We can take these techniques a step further and make use of agent training to help build the game itself. Instead of applying RL to capture player behaviors, we consider an approach to game-play design where the player agents learn behavior policies from the game designers. \blue{The primary motivation of that is to give direct control into the designer hands and enable easy interactive creation of various behavior types. At the same time, we aim to complement organic demonstrations with bootstrap and heuristics to eliminate the need for a human to train an agent on the states normally not encountered by humans, e.g., unblocking an agent using obstacle avoidance.}

\subsection{Human-Like Exploration in an Open-World Game}

To bridge the gap between the agent and the designer, we introduce imitation learning (IL) to our system~\cite{GAIL,IL1, IL2,IL3}. 
In the present application, IL allows us to translate the intentions of the game designer into a primer and a target for our agent learning system.
Learning from expert demonstrations has traditionally proved very helpful in training agents, \blue{including in games~\cite{thurau2007bayesian}}. In particular, the original Alpha Go~\cite{alpha-go} used expert demonstrations in training a deep Q network. While subsequent work argued that learning via self-play could have better asymptotic return, the performance gain comes with significantly higher training computational resource costs and superhuman performance is not seeked in this work.
Other cases preferred training agents on relatively short demonstrations played by developers or designers~\cite{oneshot-IL}.

\begin{figure*}
  \centering
  \includegraphics[width=0.77\linewidth]{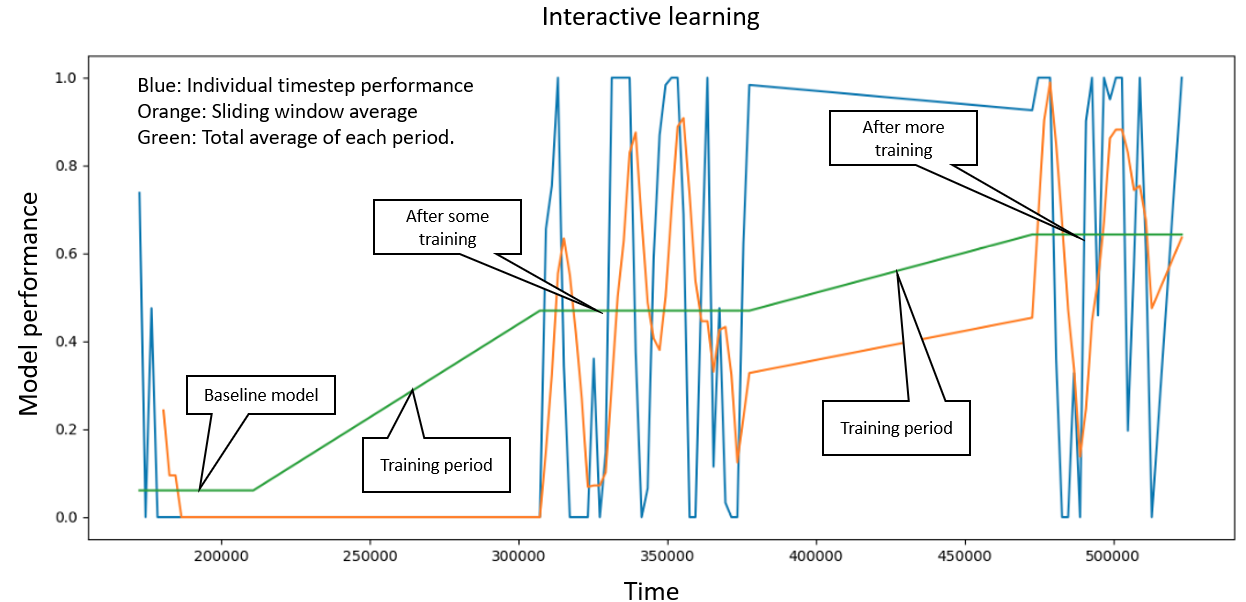}
  \vspace{-.08in}
  \caption{Model performance measures the probability of the event that the Markov agent finds at least one previous action from human-played demonstration episodes in the current game state. The goal of interactive learning is to add support for new game features to the already trained model or improve its performance in underexplored game states. Plotted is the model performance during interactive training from demonstrations in a proprietary open-world game as a function of time measured in \blue{milliseconds (with the total duration around 10 minutes).} }
  \label{Figure:interactive_training}
 \vspace{-.08in}
\end{figure*}

In this application, we consider training artificial agents in an open-world video game, where the game designer is interested in training non-player characters that \blue{exhibit} certain behavioral styles. \blue{The game we are exploring is a shooter with contextual game-play and destructible environment. We focus on single-player, which provides an environment tractable yet rich enough to test our approach. Overall the dimensionality of the agent state can grow to several dozens of continuous and categorical variables. We construct similar states for interactable NPCs. }

\blue{The NPCs in this environment represent adversarial entities, trying to attack the agent. Additionally, the environment can contain objects of interest, like ammo boxes, dropped weapons, etc. The environment itself is non-deterministic stochastic, i.e., there is no single random seed which we can set to control all random choices in the environment. Additionally, frequent saving and reloading game state is not practical due to relatively long loading times.
}

\blue{The main objective for us in this use case is to provide designer with a tool to playtest the game by interacting with the game in a number of particular styles to emulate different players. The styles can include: 
\begin{itemize}
    \item Aggressive: the agent tries to defeat adversarial NPCs,
    \item Sniper: the agent finds a good sniping spot and waits for adversaries to appear in its cone of sight to shoot them,
    \item Exploratory: the agent attempts to explore as many locations and objects of interest as possible while actively trying to avoid combat,
    \item Sneaky: the agent tries focus on its objectives while avoiding combat.
\end{itemize}
}


\blue{An agent trained this way can also be playing as an ``avatar'' of an actual human player, to stand-in for the players when they are not online or to fill a vacant spot in a squad. The agents are not designed to have any specific level of performance and they may not necessarily follow any long-term goals. These agents are intended to explore the game and also be able to interact with human players at a relatively shallow level of engagement.
In summary,} we want to efficiently train an agent using demonstrations capturing only certain elements of the game-play. The training process has to be computationally inexpensive and the agent has to imitate the behavior of the teacher(s) by mimicking their relevant style (in a statistical sense) for \emph{implicit} representation of the teacher's 
objectives. 

Casting this problem directly into the RL framework is complicated. First, it is not straightforward how to design a rewarding mechanism for imitating the style of the expert. 
While inverse RL aims at solving this problem, its applicability is not obvious given the reduced representation of the huge state-action space that we deal with and the ill-posed nature of the inverse RL problem~\cite{apprenticeship-learning,IRL}.
Second, the RL training loop often requires thousands of episodes to learn useful policies, directly translating to a high cost of training in terms of time and computational resources. \blue{Hence, rather than using more complex solutions such as generative adversarial imitation learning~\cite{GAIL} which use an RL network as their generator,} we propose a solution to the stated problem based on an ensemble of multi-resolution Markov models. \blue{One of the major benefits of the proposed model is the ability to perform an interactive training within the same episode. As useful byproduct of our formulation, we can also sketch a mechanism for numerical evaluation of the style associated with the agents we train. We outline the main elements of the approach next, for additional details refer to \cite{ICML-HILL,Igor-NeurIPS18,AAAI-Make, LLNL_CASIS_2019} }.

{
\subsubsection{Markov Decision Process with Extended State} 
We place the problem into the standard MDP framework and augment it as follows. Firstly, we ignore differences between the observation and the actual state $s$ of the environment. The actual state may be impractical to expose to the agent. 
To mitigate partial observability, we extended observations with a short history of previously taken actions. In addition to implicitly encoding the intent of a teacher and their reactions to potentially richer observations, it helps to preserve the stylistic elements of human demonstrations. 

Concretely, we assume the following. The interaction of the agent and the environment takes place at discrete moments $t=1,\dots, T$ with the value of $t$ trivially observable by the agent. After receiving an observation $s_t$ at time $t$, the agent can take an action $a_t$ from the set of allowed actions $A(s, t)$, using policy $\pi: s \to a$. Executing an action  results in a new state $s_{t+1}$. Since we focus on the stylistic elements of the agent behavior, the rewards are inconsequential for the model we build, and we drop them from the discussion. Considering the episode-based environment, a complete episode is then $E=\{(s_t, a_t)\}_{t \in {1,\dots,T}}$. The fundamental assumption regarding the described decision process is that it has the Markov property.

We also consider a recent history of the past $n$ actions, where $1 \leq  n < T$, $\alpha_{t, n} := a_{t-n}^{t-1}=\{a_{t-n}, \dots, a_{t-1}\}$, whenever it is defined in episode $E$. For $n=0$, we define $a_{t,0}$ as the empty sequence. 
We augment observed state $s_t$ with the action history $\alpha_{t,n}$, to obtain \textit{extended state} $S_{t,n}=(s_t, \alpha_{t,n})$. 

The purpose of including the action history is to capture additional information from human input during interactive demonstrations. An extended policy $\pi_{n}$, which operates on the extended states $\pi_{n} : S_{t,n} \to a_t$, is useful for modeling human actions in a manner similar to $n$-grams text models in natural language processing (NLP) (e.g., \cite{KaminskiMP}, \cite{davidwrite}, \cite{andresen2017approximating}). Of course, the analogy with $n$-gram models in NLP works only if both state and action spaces are discrete. We address this restriction in the next subsection using multi-resolution quantization. 

For a discrete state-action space and various $n$, we can compute probabilities $P\{a_t|S_{t,n}\}$ of transitions $S_{t,n} \to a_t$ occurring in demonstrations and use them as a Markov model $M_n$ of order $n$ actions. We say that the model $M_n$ is defined on an extended state $S_{.,n}$ if the demonstrations contain at least one occurrence of $S_{.,n}$. When a model $M_n$ is defined on $S$, we can use $P\{a_t|S_{t,n}\}$ to sample the next action from all ever observed next actions in state $S_{.,n}$. Hence, $M_n$ defines a partial stochastic mapping $M_n: S_{.,n} \to A$ from extended states to action space $A$.

\subsubsection{Stacked Markov models} 
We call a sequence of Markov models $\mathcal{M}_n = \{M_i\}_{i=0,\dots,n}$ a stack of models. A (partial) policy defined by $\mathcal{M}_n$ computes the next action at a state $s_t$, see \cite{ICML-HILL} for the pseudo-code of the corresponding algorithm. Such policy performs a simple behavior cloning. The policy is partial since it may not be defined on all possible extended states and needs a fallback policy $\pi_*$ to provide a functional agent acting in the environment. 

Note that it is possible to implement sampling from a Markov model using an $\mathcal{O}(1)$ complexity operation with hash tables, making the inference very efficient and suitable for real-time execution in a video game.

\subsubsection{Quantization} 
Quantization (aka discretization) works around the limitation of discrete state-action space, enabling the application of the Markov Ensemble approach to environments with continuous dimensions. Quantization is commonly used in solving MDPs \cite{RL_state_of_the_art} and has been extensively studied in the signal processing literature \cite{digital_signal_proc}, \cite{vect_quant}. Quantization schemes that have been optimized for specific objectives can lead to significant gains in model performance, improving various metrics vs. ad-hoc quantization schemes  \cite{RL_state_of_the_art}, \cite{Pages}. 

Instead of trying to pose and solve the problem of optimal quantization, we use a set of quantizers covering a range of schemes from coarse to fine. At the conceptual level, such an approach is similar to multi-resolution methods in image processing, mip-mapping and Level-of-Detail (LoD) representations in computer graphics \cite{comp_graph}. The simplest quantization is a uniform one with bin size $\sigma$, i.e.,  $Q_{\sigma}(x) = \sigma \floor*{\frac{x}{\sigma}}.$
For each continuous variable in the state-action space, we consider a sequence of quantizers with decreasing step size $Q = \{Q_{\sigma_j}\}_ {j=0,\dots, K}$, $\sigma_j > \sigma_{j+1}$, which naturally gives a quantization sequence $\Bar{Q}$ for the entire state-action space, provided $K$ is fixed across the continuous dimensions. To simplify notation, we collapse the sub index and write $Q_{j}$ to stand for $Q_{\sigma_j}$. For more general quantization schemes, the main requirement is the decreasingly smaller reconstruction error for $Q_{j+1}$ in comparison to $Q_j$.

For an episode $E$, we compute its quantized representation in an obvious component-wise manner: 
\begin{equation} \label{eq:quantization}
E_j = \Bar{Q}_j(E)=\{(\Bar{Q}_j(s_t), \Bar{Q}_j(a_t))\}_{t \in {1,\dots,T}}    
\end{equation}
which defines a multi-resolution representation of the episode as a corresponding ordered set $\{E_j\}_{j \in \{0, \dots, K\}}$ of quantized episodes, where $\Bar{Q}$ is the vector version of quantization $Q$.

In the quantized Markov model $M_{n,j}= \Bar{Q}_j(M_n)$, constructed from the episode $E_j$, we compute extended states using the corresponding quantized values. The extended state is $\Bar{Q}_j(S_{t,n})=(\Bar{Q}(s_t), \Bar{Q}(\alpha_{t,n}))$. Further, we define the model $\Bar{Q}_j(M_n)$ to contain probabilities $P\{a_t|\Bar{Q}_j(S_{t,n})\}$ for the \emph{original} action values. In other words, we do not rely on the reconstruction mapping $\Bar{Q}_j^{-1}$ to recover action, but store the original actions explicitly. Continuous action values tend to be unique and the model samples from the set of values observed after the occurrences of the corresponding extended state. Our experiments show that replaying the original actions instead of their quantized representation provides better continuity and natural true-to-the-demonstration look of the cloned behavior.

\subsubsection{Markov Ensemble} 
Combining stacking and multi-resolution quantization of Markov models, we obtain Markov Ensemble $\mathcal{E}$ as an array of Markov models parameterized by the model order $n$ and the quantization schema $Q_{j}$.
Note, that with the coarsest quantization $\sigma_0$ present in the multi-resolution schema, the policy should always return an action sampled using one of the quantized models, which at the level $0$ always finds a match. Hence, such models always ``generalize'' by resorting to simple sampling of actions when no better match found in the observations. Excluding too coarse quantizers and Markov order 0 will result in executing some ``default policy'' $\pi_*$, which we discuss in the next section. The agent execution with the outlined ensemble of quantized stacked Markov models is easy to express as an algorithm, which in essence boils down to a look-up table \cite{ICML-HILL}.

\subsubsection{Interactive Training of Markov Ensemble with human in the loop (HITL) learning} 
If the environment allows a human to override the current policy and record new actions, then we can generate demonstrations produced interactively. For each demonstration, we construct a new Markov Ensemble and add it to the sequence of existing models. The policy based on these models consults with the latest one first. If the model fails to produce an action, the next model is asked, until there are no other models. Thanks to the sequential organization, the latest demonstrations take precedence of the earlier ones, allowing correcting previous mistakes or adding new behavior for the previously unobserved situations. We illustrate the logic of such an interaction with the sample git repository \cite{ibor_github_jun2019}. 
In our case studies, we show that often even a small number of strategically provided demonstrations results in a well-behaving policy.

While the spirit of the outlined idea is similar to that of DAgger~\cite{dagger}, 
providing corrective labels on the newly generated samples is more time consuming than providing new demonstration data in under-explored  states.
The interactivity could also be used to support newly added features or to update the existing model otherwise. The designer can directly interact with the game, select a particular moment where a new human demonstration is required and collect new demonstration data without reloading the game. The interactivity eliminates most of the complexity of the agent design process and brings down the cost of gathering data from under-explored parts of the state space.

We report an example chart for such an interactive training in Fig.~\ref{Figure:interactive_training}. 
The goal in this example is to train an agent capable of an attack behavior. The training on the figure starts with the most basic game-play when the designer provides a demonstration for approaching the target. The next training period happens after observing the trained model for a short period of time. In between the training training periods, the designer makes sure that the agent reaches the intended state and is capable of executing already learned actions. The second training period adds more elements to the behavior. The figure covers several minutes of game-play and the sliding window size is approximately one second, or 30 frames. The competence here is equated to the model performance and is a metric of how many states the model can handle by returning an action. The figure shows that the model competence grows as it accumulates more demonstrations in each of the two training segments. The confidence metric is a natural proxy for evaluating how close stylistically is the model behavior to the demonstrations. Additional details on the interactive training are available from \cite{ICML-HILL} and the repository \cite{ibor_github_jun2019} which allows experimentation with two classic control OpenAI environments.

The prolonged period of training may increase the size of the model with many older demonstrations already irrelevant, but still contributing to the model size. Instead of using rule-based compression of the resulting model ensemble, in the next subsection, we discuss the creation of a DNN model trained from the ensemble of Markov models via a novel bootstrap approach using the game itself as the way to compress the model representation and strip off obsolete demonstration data. Using the proposed approach, we train an agent that satisfies the design needs in only a few hours of interactive training.

\subsubsection{A sketch of style distance with Markov Ensemble}
The models defined above allows us to introduce a candidate metric for measuring stylistic difference between behaviors $V$ and $W$ represented by the corresponding set of episodes. For a fixed quantization scheme, we can compute a sample distribution of the $n$-grams for both behaviors, which we denote as $v_n$ and $w_n$. Then the ``style'' distance $D=D_{\lambda, N}(V, W)$ between $V$ and $W$ can be estimated using the formula:
$$D(V, W) = \frac{\lambda}{1 - \lambda}  \sum_{n=0}^{N} \lambda^{n} d(v_n, w_n) + \frac{\lambda^{N+1}}{1 - \lambda} d(v_N, w_N),$$
where $\lambda \in (0,1)$ controls the contributions of different $n$-gram  models. As defined, larger $\lambda$ puts more weight on longer $n$-grams and as such values more complex sequence of actions more. The function $d$ is one of the probability distances. We used Jensen-Shannon (JSD) and Hellinger (HD); both in the range $[0,1]$, hence for all $V$ and $W$, $D \in [0,1]$. The introduced distance can augment the traditional RL rewards to preserve style during training without human inputs as we discuss in \cite{LLNL_CASIS_2019}. However, the main motivation of introducing distance $D$ is to provide a numerical metric to evaluate how demonstrations and the learned policy differ in terms of style without visually inspecting them in the environment.

}

\subsection{Bootstrapped DNN agent}

\blue{The ensemble of multi-resolution Markov models described in the previous section suffers from several drawbacks. One is the linear growth of the model size with demonstration data. The other problem stems from the limited nature of the human demonstrations. In particular, humans proactively take certain actions and there are only few if any ``negative'' examples where humans fails to navigate smoothly. 
Imitation learning is well known to suffer from propagation of errors reaching states that are not represented in the demonstration data.
In these situations, the Markov agent may get blocked and act erratically. To address both of these issues, we introduce a bootstrapped DNN agent.}

\blue{When generating boostrapped episodes, we use the Markov agent augmented with some common sense rules (heuristics addressing the states not encountered in demonstrations). 
For instance, for a blocked state, a simple obstacle avoidance fall-back policy can help the agent unblock. Combining such rules with the supervised learning from human demonstrations allows to make the boostrapped training dataset much richer. }

We treat the existing demonstrations as a training set for a supervised policy, which predicts the next action conditioned on the gameplay history, i.e., a sequence of observed state-action pairs. 
Since our demonstration dataset is relatively small, we generate more data (bootstrap) by letting the trained Markov agent interact with the game environment.

{
\renewcommand{\arraystretch}{1.3}
\begin{table}[t]
  \caption{Comparison between OpenAI 1v1 Dota 2 Bot \cite{openai-dota2} training metrics and training an agent via bootstrap. \blue{The comparison is not 1-to-1 because the training objectives are very different. However, the environments are similar in complexity. 
 These metrics highlight the practical training of agents during the game development cycle. The point is to illustrate that the training objectives play a critical role.}}
  \label{stats-table}
  \centering
  \vspace{0.12in}
  \begin{tabular}{l|l|l}
    & OpenAI & Bootstrapped \\
    & 1v1 Bot & Agent \\
    \hline
    Experience & $\sim$300 years & $\sim$5 min \\
               & (per day) & demonstrations\\
    \hline
    Bootstrap using &  N/A & $\times$5-20 \\
    game client &   & \\
    \hline
    CPU & 60,000 CPU & 1 local CPU \\
        & cores on Azure &  \\
    \hline
    GPU & 256 K80 GPUs & N/A \\
       & on Azure &  \\
    \hline
    Size of& $\sim$3.3kB & $\sim$0.5kB\\ 
    observation & & \\ 
    \hline
    Observations  & 10 & 33 \\
    per second & & \\
  \end{tabular}
\end{table}
}

Such a bootstrap process is easy to parallelize in offline policy training as we can have multiple copies of the agent running without the need to cross-interact, as opposed to online algorithms such as A3C \cite{async_rl}. The generated augmented dataset is used to train a more sophisticated policy (deep neural network). Due to partial observability, the low dimensionality of the feature space results in fast training in a wide range of model architectures, allowing a quick experimentation loop. We converged on a simple model with a single ``wide'' hidden layer for motion control channels and a fully-connected DNN for discrete channels responsible for turning on/off actions like sprint, firing, climbing. The approach shows promise even with many yet unexplored opportunities to improve its efficiency. 

\blue{A reasonable architecture for both DNN models can be inferred from the tasks they solve. For the motion controller, the only hidden layer roughly corresponds to the temporal-spacial quantization levels in the base Markov model. When using ReLUs for motion controller hidden layer, we start experimentation with their number equal to the double number of the quantization steps per input variable. Intuitively, training encodes those quantization levels into the layer weights. Adding more depth may help to better capture stylistic elements of the motion. To prevent overfitting, the number of model parameters  should be chosen based on the size of the training dataset. In our case, overfitting to the few demonstrations may result in better representation of the style, yet may lead to the degraded in-game performance, e.g., an agent will not achieve game-play objectives as efficiently. In our experiments, we find that consistent (vs. random) demonstrations require only single hidden layer for the motion controller to reproduce basic stylistic features of the agent motion. A useful rule of thumb for discrete actions DNN is to start with the number of layers roughly equal to the maximum order of Markov model used to drive the bootstrap and conservatively increase the model complexity only as needed. For such a DNN, we are using fully connected layers with the number of ReLUs per layer roughly equal to doubled the dimensionality of the input space. We also observed that recurrent networks do not provide much improvement perhaps because the engineered features capture the essence of the game state.
}


Table \ref{stats-table} illustrates the computational resources required by this approach as compared to training 1v1 agents in Dota 2~\cite{openai-dota2}. While we acknowledge that the goal of our agent is not to play optimally against the opponent, we observe that using model-based training augmented with expert demonstrations to solve the Markov decision process, in a complex game, results in huge computational savings compared to an optimal reinforcement learning approach.

\subsection{\blue{Cooperative (Assistive) game-playing AI}}
\label{sec:team-sports}

\blue{Our last case study involves a team sports game, where the designer's goal is to train agents that can learn strategic teamplay to complement arbitrary human player styles. For example, if the human player is more offensive, we would like their teammate agent to be more defensive and vice versa. The game in question involves two teams trying to score the most points before time runs out. To score a point, the team needs to put the ball past the goal line on their opponent's side of the field. Similar to several team sports games, the players have to fight for ball possession for them to be able to score, and hence ball control is a big component of this game.
This is a more complex challenge compared to the previous case study that concerned exploration of a game world. As the agent in this game is required to make strategic decisions, we resort to reinforcement learning.}

\begin{figure}
    \centering
    \includegraphics[width=0.4\linewidth, angle=90]{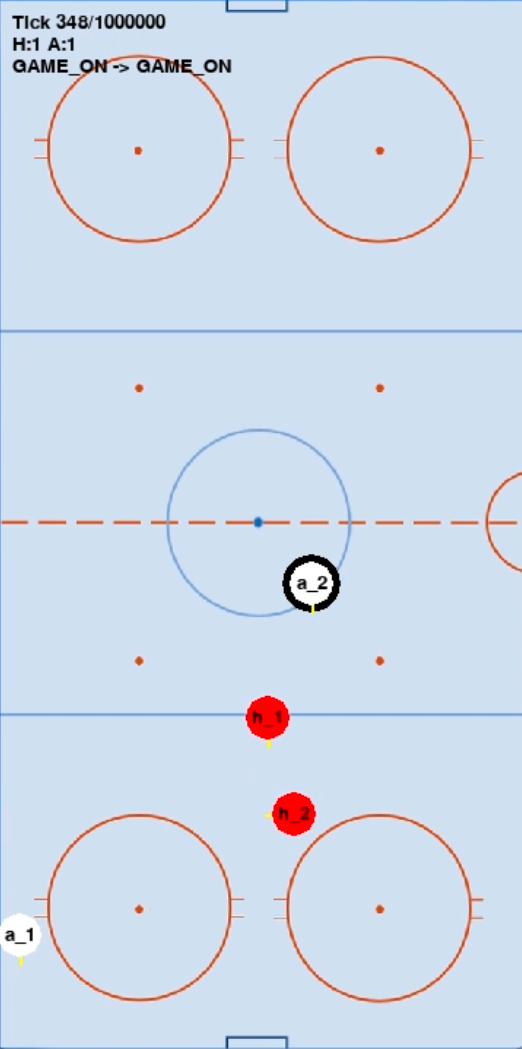}
    \caption{A screen shot of the simple team sports simulator (STS2). Red agents are home agents attempting to score at the left end and white agents are away agents attempting to score at the right end. The highlighted player has possession of the ball and the arrows demonstrate a pass/shoot attempt.}\label{fig:STS2} 
  \end{figure}

Our training takes place on simple team sports simulator (STS2).\footnote{We intend to release this environment as an open-source package.} A screenshot of STS2 game-play is shown in Fig.~\ref{fig:STS2}. The simulator embeds the rules of the game and the physics at a high level. Each of the players can be controlled by a human, a pre-built rule-based agent, or any other learned policy.
The behavior of a rule-based agent is controlled by a handful of rules and constraints that govern their game-play strategy, and is most similar to game controlled opponents usually implemented in adversarial games.
The STS2 state space consists of player coordinates, their velocities and an indicator of ball possession. The action space is discrete and is considered to be left, right, forward, backward, pass, and shoot. Although the player can hit two or more of the actions together, we exclude that possibility to keep the complexity of the action space manageable.

{
Our goal is to train a teammate agent that can adapt to a human player's style. As the simplest multi-agent mode, we consider a 2v2 game.
We show two scenarios. In the first case, we train an agent to cooperate with a novice human player. In the second case, we train a defensive agent that complements a high-skill offensive agent teammate. A more comprehensive study on the material in this section appears in~\cite{MAL-team-sports}.

\subsubsection{Game-playing AI to assist a low-skill player}
We consider training an agent in a 2v2 game \blue{that can assist a low-skill player.} 
We let rule-based agents take control of the opponent players. We also choose a low-skill rule-based agent to control the teammate player. The goal is to train a cooperative agent that complements the low-skill agent. In this experiment, 
we provide a `+/-1' reward for scoring. We also provide  a `+/-0.8` {\em individual} reward for the agent for  gaining/losing the possession of the ball. This reward promotes the agent to gain the ball back from the opponent and score.
We ran the experiment using DQN, PPO, and Rainbow (with its default hyperparameters). PPO requires an order of magnitude less trajectories for convergence, and the final policy is similar to that of DQN. However, Rainbow did not converge at all with the default hyperparameters and we suspect that the prioritized experience replay~\cite{prioritized-experience-replay} is sensitive to hyperparameters.

The team statistics for this agent are shown in Table~\ref{tab:1w1v2_offensive}. As can be seen, the agent has learned an offensive game-play style where it scores most of the time. It also keeps more possession than the rest of the agents in the game.

{
\renewcommand{\arraystretch}{1.3}
\begin{table}[h]
  \caption{Offensive DQN agent in a 2v2 partnered with the rule-based agent versus two rule-based agents. Rewards: sparse `+/-1' for scoring and individual `+/-0.8' for win/lose possession of the ball. 
  }
  \label{tab:1w1v2_offensive}
  \centering
  \begin{tabular}{l|c|c|c|c}
    Statistics & DQN-1 & Rule-based Agent & Opponent 1 & Opponent 2\\

    \hline
    Score rate & 54\% & 20\% & 13\% & 13\%\\
    \hline
    Possession & 30\% & 18\%  & 26\% & 26\% \\
  \end{tabular}
\end{table}
}

\subsubsection{Game-playing AI to assist a high-skill offensive player}
Next, we report training an agent that complements a high-skill offensive player.
In particular, we train an agent that complements the DQN-1 that was trained in the previous experiment. 
We train another agent as the teammate using exactly the same rewarding mechanism as the one used in training the offensive DQN-1 agent. 
The statistics of the game-play for the two agents playing together against the rule-based agent are shown in Table~\ref{tab:2v2_offensive}. While the second agent is trained with the same reward function as the first one, it is trained in a different environment as it is partnered with the previous offensive DQN-1 agent. As can be seen, the second agent now becomes defensive and is more interested in protecting the net, recovering ball possession, and passing it to the offensive teammate. We can also see that the game stats for DQN-2 are similar to that of the rule-based agent in the previous experiment.

{
\renewcommand{\arraystretch}{1.3}
\begin{table}[h]
  \caption{Two DQN agents in a 2v2 match against two rule-based agents, with a sparse `+/-1' reward for scoring and a `+/0.8' individual reward for gaining/losing the possession of the ball. 
  }
  \label{tab:2v2_offensive}
  \centering
  \begin{tabular}{l|c|c|c|c}
    Statistics & DQN-1& DQN-2 & Opponent 1 & Opponent 2\\

    \hline
    Score rate & 50\% & 26\% & 12\% & 12\%\\
    \hline
    Possession & 28\% & 22\%  & 25\% & 25\% \\
  \end{tabular}
\end{table}
}

}

We repeated these experiments using PPO and Rainbow as well. We observe that the PPO agent's policy converges quickly to a simple one. When it is in possession of the ball, it wanders around in its own half without attempting to cross the half-line or to shoot until the game times out. This happens because the rule-based agent is programmed not to chase the opponent in their half when the opponent has the ball, and hence, the game goes on as described until timeout with no scoring on either side. PPO has clearly reached a local minimum in the space of policies, which is not unexpected as it is optimizing the policy directly.
Finally, the Rainbow agent does not learn a useful policy in this case.

\section{Concluding Remarks}
\label{sec:conclusion}

In this paper, we presented our efforts to create intelligent agents that can assist game designers in building games. To this end, we outlined a training pipeline, designed to train agents in games. We presented four case studies, two on creating playtesting agents and two on creating game-playing agents. Each use case showcased intelligent agents that strike a balance between skill and style. 

In the first case study, we considered The Sims Mobile in its early development stage. We showed that the game dynamics could be fully extracted in a lightweight model of the game. This removed the need for learning; and the game-play experience was modeled using much simpler planning methods. The playtesting agent modeled with the A* algorithm proved effective because of the straightforward skill requirement, i.e., fast progression with minimum number of actions taken.


In the second case study, we considered a mobile game with large state and action spaces, where the designer's objective was to measure an average player's progression. We showed how model-free RL could  inform the designer of their design choices for the game.
The game presented a challenge in which the player's choice of actions on resource management would manifest itself near the end of the game in interactions with other players (creating an environment with delayed rewards). Our experiments demonstrated that the choice of the observation space could dramatically impact the effectiveness of the solutions trained using deep RL.
We are currently investigating proper reward shaping schemes as part of a hierarchical gameplay solution for this game.

In the third case study, we considered an open-world HD game with the goal of imitating gameplay demonstrations provided by the game designer. We used a multi-resolution ensemble of Markov models as a baseline in this environment. 
While the baseline model performed well in most settings, it encountered poor generalization in underexplored states. 
We addressed this challenge on three fronts: basic rules, human-in-the-loop learning, and compressing the ensemble into a compact representation.
We augmented the model with simple rules to avoid unintended states not present in human demonstrations. 
The end-to-end training of the baseline, taking only a few hours, allowed us to quickly iterate with the game designer in a human-in-the-loop setting to correct any unintended behavior.
Finally, we bootsrapped a supervised DNN model using the ensemble model as a simulator to generate training data, resulting in a compressed model with fast inference and better generalization.

In the last case study, we considered a team sports game, where the goal was to train game-playing agents 
that could complement human players with different skills to win against a competent opposing team. 
In addition to the reward function, the emergent behavior of an agent trained using deep RL was also impacted by the style of their teammate player.
This made reward shaping extremely challenging in this setting.
As part of this investigation, we also observed that the state-of-the-art open-source deep RL models are heavily tuned to perform well on benchmark environments including Atari games. 
We are currently investigating meta-policies that could adapt to a variety of teammates and opponents without much tuning.

These four case studies presented in this work highlight the challenges faced by game designers in training intelligent agents for the purpose playtesting and gameplaying AI. We would like to share two main takeaways learned throughout this work as guiding principles for the community: (1) depending on the problem at hand, we need to resort to a variety of techniques, ranging from planning to deep RL, to effectively accomplish the objectives of designers; (2) the learning potential of state-of-the-art deep RL models does not seamlessly transfer from the benchmark environments to target ones without heavily tuning their hyperparameters, leading to linear scaling of the engineering efforts and computational cost with the number of target domains.


\section*{Acknowledgement}
The authors are thankful to EA Sports and other game team partners for their support and collaboration. \blue{The authors also would like to thank the anonymous reviewers and the EIC for their constructive feedback.}

\bibliographystyle{IEEEtran}  
\bibliography{MAS}  

\begin{thebibliography}{10}
\providecommand{\url}[1]{#1}
\csname url@samestyle\endcsname
\providecommand{\newblock}{\relax}
\providecommand{\bibinfo}[2]{#2}
\providecommand{\BIBentrySTDinterwordspacing}{\spaceskip=0pt\relax}
\providecommand{\BIBentryALTinterwordstretchfactor}{4}
\providecommand{\BIBentryALTinterwordspacing}{\spaceskip=\fontdimen2\font plus
\BIBentryALTinterwordstretchfactor\fontdimen3\font minus
  \fontdimen4\font\relax}
\providecommand{\BIBforeignlanguage}[2]{{%
\expandafter\ifx\csname l@#1\endcsname\relax
\typeout{** WARNING: IEEEtran.bst: No hyphenation pattern has been}%
\typeout{** loaded for the language `#1'. Using the pattern for}%
\typeout{** the default language instead.}%
\else
\language=\csname l@#1\endcsname
\fi
#2}}
\providecommand{\BIBdecl}{\relax}
\BIBdecl

\bibitem{Fernando-A-star}
F.~D.~M. Silva, I.~Borovikov, J.~Kolen, N.~Aghdaie, and K.~Zaman, ``Exploring
  gameplay with {AI} agents,'' in \emph{AIIDE}, 2018.

\bibitem{Igor-NeurIPS18}
\BIBentryALTinterwordspacing
I.~Borovikov and A.~Beirami, ``Imitation learning via bootstrapped
  demonstrations in an open-world video game,'' in \emph{NeurIPS 2018 Workshop
  on Reinforcement Learning under Partial Observability}, Dec 2018. [Online].
  Available:
  \url{https://www.ias.informatik.tu-darmstadt.de/uploads/Team/JoniPajarinen/RLPO2018_paper_17.pdf}
\BIBentrySTDinterwordspacing

\bibitem{Yunqi-NeurIPS18}
\BIBentryALTinterwordspacing
Y.~Zhao, A.~Beirami, M.~Sardari, N.~Aghdaie, and K.~Zaman, ``Training agents to
  play modern games: Challenges and opportunities,'' in \emph{NeurIPS 2018
  Workshop on Reinforcement Learning under Partial Observability}, Dec 2018.
  [Online]. Available:
  \url{https://www.ias.informatik.tu-darmstadt.de/uploads/Team/JoniPajarinen/RLPO2018_paper_19.pdf}
\BIBentrySTDinterwordspacing

\bibitem{AAAI19-RLG}
\BIBentryALTinterwordspacing
I.~Borovikov, Y.~Zhao, A.~Beirami, J.~Harder, J.~Kolen, J.~Pestrak, J.~Pinto,
  R.~Pourabolghasem \emph{et~al.}, ``Winning isn’t everything: Training
  agents to playtest modern games,'' in \emph{AAAI Workshop on Reinforcement
  Learning in Games}, Jan 2019. [Online]. Available:
  \url{http://aaai-rlg.mlanctot.info/papers/AAAI19-RLG-Paper36.pdf}
\BIBentrySTDinterwordspacing

\bibitem{AAAI-Make}
\BIBentryALTinterwordspacing
I.~Borovikov and A.~Beirami, ``From demonstrations and knowledge engineering to
  a {DNN} agent in a modern open-world video game,'' in \emph{AAAI 2019 Spring
  Symposium on Combining Machine Learning with Knowledge Engineering}, Mar
  2019. [Online]. Available:
  \url{https://proceedings.aaai-make.info/short2.pdf}
\BIBentrySTDinterwordspacing

\bibitem{ICML-HILL}
I.~Borovikov, J.~Harder, M.~Sadovsky, and A.~Beirami, ``Towards interactive
  training of non-player characters in video games,'' \emph{arXiv preprint
  arXiv:1906.00535}, 2019.

\bibitem{MAL-team-sports}
\BIBentryALTinterwordspacing
Y.~Zhao, I.~Borovikov, J.~Rupert, C.~Somers, and A.~Beirami, ``On multi-agent
  learning in team sports games,'' in \emph{ICML 2019 Workshop on Imitation,
  Intent, and Interaction (I3)}, June 2019. [Online]. Available:
  \url{https://drive.google.com/drive/folders/1rgx4G0Jl6XG20AgpoDJwnKbwBt7Ei1SJ}
\BIBentrySTDinterwordspacing

\bibitem{LLNL_CASIS_2019}
\BIBentryALTinterwordspacing
I.~Borovikov, J.~Harder, M.~Sadovsky, and A.~Beirami, ``Towards a
  representative metric of behavior style in imitation and reinforcement
  learning,'' in \emph{The 23rd Annual Signal and Image Sciences Workshop at
  Lawrence Livermore National Laboratory, Center for Advanced Signal Image
  Sciences (CASIS)}, May 2019. [Online]. Available:
  \url{https://casis.llnl.gov/content/pages/casis-2019/docs/Borovikov_CASIS_2019_NO-LLNL-IM.pdf}
\BIBentrySTDinterwordspacing

\bibitem{ibor_github_jun2019}
I.~Borovikov and J.~Harder, ``{Interactive Training (code base)},''
  \url{https://github.com/electronicarts/interactive_training}, 2019.

\bibitem{deep-blue}
M.~Campbell, A.~J. Hoane~Jr, and F.-h. Hsu, ``Deep {Blue},'' \emph{Artificial
  intelligence}, vol. 134, no. 1-2, pp. 57--83, 2002.

\bibitem{alpha-go}
D.~Silver, A.~Huang, C.~J. Maddison, A.~Guez, L.~Sifre, G.~Van Den~Driessche,
  J.~Schrittwieser, I.~Antonoglou, V.~Panneershelvam, M.~Lanctot \emph{et~al.},
  ``{Mastering the game of Go with deep neural networks and tree search},''
  \emph{Nature}, vol. 529, no. 7587, pp. 484--489, 2016.

\bibitem{AlphaStar}
{AlphaStar}, 2019, [Online, January
  2019]~\url{https://deepmind.com/blog/alphastar-mastering-real-time-strategy-game-starcraft-ii}.

\bibitem{drachen2009playerstyle}
A.~Drachen, A.~Canossa, and G.~N. Yannakakis, ``Player modeling using
  self-organization in tomb raider: Underworld,'' in \emph{2009 IEEE symposium
  on computational intelligence and games}.\hskip 1em plus 0.5em minus
  0.4em\relax IEEE, 2009, pp. 1--8.

\bibitem{gow2012unsupervisedplayerstyle}
J.~Gow, R.~Baumgarten, P.~Cairns, S.~Colton, and P.~Miller, ``Unsupervised
  modeling of player style with lda,'' \emph{IEEE Transactions on Computational
  Intelligence and AI in Games}, vol.~4, no.~3, pp. 152--166, 2012.

\bibitem{ortega2013imitating}
J.~Ortega, N.~Shaker, J.~Togelius, and G.~N. Yannakakis, ``Imitating human
  playing styles in super mario bros,'' \emph{Entertainment Computing}, vol.~4,
  no.~2, pp. 93--104, 2013.

\bibitem{fun-in-games}
M.~A. Federoff, ``Heuristics and usability guidelines for the creation and
  evaluation of fun in video games,'' Ph.D. dissertation, Citeseer, 2002.

\bibitem{immersion-in-games}
P.~Cairns, A.~Cox, and A.~I. Nordin, ``Immersion in digital games: review of
  gaming experience research,'' \emph{Handbook of digital games}, vol.~1, p.
  767, 2014.

\bibitem{de2017contemporaryboardgameai}
F.~De~Mesentier~Silva, S.~Lee, J.~Togelius, and A.~Nealen, ``{AI}-based
  playtesting of contemporary board games,'' in \emph{Foundations of Digital
  Games 2017}.\hskip 1em plus 0.5em minus 0.4em\relax ACM, 2017.

\bibitem{hom2007automatic}
V.~Hom and J.~Marks, ``Automatic design of balanced board games,'' in
  \emph{Proceedings of the AAAI Conference on Artificial Intelligence and
  Interactive Digital Entertainment (AIIDE)}, 2007, pp. 25--30.

\bibitem{r2014}
J.~Krucher, ``Algorithmically balancing a collectible card game,'' Bachelor's
  Thesis, ETH Zurich, 2015.

\bibitem{mahlmann2012evolving}
T.~Mahlmann, J.~Togelius, and G.~N. Yannakakis, ``Evolving card sets towards
  balancing dominion,'' in \emph{Evolutionary Computation (CEC), 2012 IEEE
  Congress on}.\hskip 1em plus 0.5em minus 0.4em\relax IEEE, 2012, pp. 1--8.

\bibitem{silva2019evolving}
F.~de~Mesentier~Silva, R.~Canaan, S.~Lee, M.~C. Fontaine, J.~Togelius, and
  A.~K. Hoover, ``Evolving the hearthstone meta,'' in \emph{IEEE Conference on
  Games}, 2019.

\bibitem{holmgard}
C.~Holmgard, M.~C. Green, A.~Liapis, and J.~Togelius, ``Automated playtesting
  with procedural personas with evolved heuristics,'' \emph{IEEE Transactions
  on Games}, 2018.

\bibitem{mugrai2019automated}
L.~Mugrai, F.~de~Mesentier~Silva, C.~Holmg{\aa}rd, and J.~Togelius, ``Automated
  playtesting of matching tile games,'' in \emph{IEEE Conference on Games},
  2019.

\bibitem{guerrero2018using}
C.~Guerrero-Romero, S.~M. Lucas, and D.~Perez-Liebana, ``Using a team of
  general ai algorithms to assist game design and testing,'' in \emph{2018 IEEE
  Conference on Computational Intelligence and Games (CIG)}.\hskip 1em plus
  0.5em minus 0.4em\relax IEEE, 2018, pp. 1--8.

\bibitem{yannakakis2014mixed}
G.~N. Yannakakis, A.~Liapis, and C.~Alexopoulos, ``Mixed-initiative
  co-creativity,'' in \emph{Proceedings of the 9th Conference on the
  Foundations of Digital Games}, 2014.

\bibitem{liapis2013sentient}
A.~Liapis, G.~N. Yannakakis, and J.~Togelius, ``Sentient sketchbook:
  Computer-aided game level authoring.'' in \emph{FDG}, 2013, pp. 213--220.

\bibitem{smith2010tanagra}
G.~Smith, J.~Whitehead, and M.~Mateas, ``Tanagra: A mixed-initiative level
  design tool,'' in \emph{Proceedings of the Fifth International Conference on
  the Foundations of Digital Games}.\hskip 1em plus 0.5em minus 0.4em\relax
  ACM, 2010, pp. 209--216.

\bibitem{shaker2013ropossum}
N.~Shaker, M.~Shaker, and J.~Togelius, ``Ropossum: An authoring tool for
  designing, optimizing and solving cut the rope levels.'' in \emph{AIIDE},
  2013.

\bibitem{search-based-generation}
J.~Togelius, G.~N. Yannakakis, K.~O. Stanley, and C.~Browne, ``Search-based
  procedural content generation: A taxonomy and survey,'' \emph{IEEE
  Transactions on Computational Intelligence and AI in Games}, vol.~3, no.~3,
  pp. 172--186, 2011.

\bibitem{pcgml}
A.~Summerville, S.~Snodgrass, M.~Guzdial, C.~Holmg{\aa}rd, A.~K. Hoover,
  A.~Isaksen, A.~Nealen, and J.~Togelius, ``Procedural content generation via
  machine learning ({PCGML}),'' \emph{IEEE Transactions on Games}, vol.~10,
  no.~3, pp. 257--270, 2018.

\bibitem{yannakakis2018artificial}
G.~N. Yannakakis and J.~Togelius, \emph{Artificial intelligence and
  games}.\hskip 1em plus 0.5em minus 0.4em\relax Springer, 2018, vol.~2.

\bibitem{samuel-checkers}
A.~Samuel, ``July 1959.“,'' \emph{Some Studies in Machine Learning Using the
  Game of Checkers.” IBM Journal of Research and Development}, vol.~3, no.~3,
  pp. 210--29.

\bibitem{MCTS}
R.~Coulom, ``{Efficient selectivity and backup operators in Monte-Carlo tree
  search},'' in \emph{International conference on computers and games}.\hskip
  1em plus 0.5em minus 0.4em\relax Springer, 2006, pp. 72--83.

\bibitem{UCT}
L.~Kocsis and C.~Szepesv{\'a}ri, ``{Bandit based Monte-Carlo planning},'' in
  \emph{European conference on machine learning}.\hskip 1em plus 0.5em minus
  0.4em\relax Springer, 2006, pp. 282--293.

\bibitem{settlersszita2009monte}
I.~Szita, G.~Chaslot, and P.~Spronck, ``Monte-carlo tree search in settlers of
  catan,'' in \emph{Advances in Computer Games}.\hskip 1em plus 0.5em minus
  0.4em\relax Springer, 2009, pp. 21--32.

\bibitem{settlerschaslot2008monte}
G.~Chaslot, S.~Bakkes, I.~Szita, and P.~Spronck, ``Monte-carlo tree search: A
  new framework for game ai.'' in \emph{AIIDE}, 2008.

\bibitem{Carcassonneheyden2009implementing}
C.~Heyden, ``Implementing a computer player for carcassonne,'' Ph.D.
  dissertation, Maastricht University, 2009.

\bibitem{7wonders}
D.~Robilliard, C.~Fonlupt, and F.~Teytaud, ``Monte-carlo tree search for the
  game of “7 wonders”,'' in \emph{Computer Games}.\hskip 1em plus 0.5em
  minus 0.4em\relax Springer, 2014, pp. 64--77.

\bibitem{MCTSTicketToRide}
C.~Huchler, ``An mcts agent for ticket to ride,'' Master's Thesis, Maastricht
  University, 2015.

\bibitem{baier2018emulating}
H.~Baier, A.~Sattaur, E.~Powley, S.~Devlin, J.~Rollason, and P.~Cowling,
  ``Emulating human play in a leading mobile card game,'' \emph{IEEE
  Transactions on Games}, 2018.

\bibitem{TDGammon}
G.~Tesauro, ``Temporal difference learning and td-gammon,''
  \emph{Communications of the ACM}, vol.~38, no.~3, pp. 58--69, 1995.

\bibitem{openai-compute}
{AI \& Compute}, 2018, [Online, May
  2018]~\url{https://blog.openai.com/ai-and-compute}.

\bibitem{DQN}
V.~Mnih, K.~Kavukcuoglu, D.~Silver, A.~A. Rusu, J.~Veness, M.~G. Bellemare,
  A.~Graves, M.~Riedmiller, A.~K. Fidjeland, G.~Ostrovski \emph{et~al.},
  ``Human-level control through deep reinforcement learning,'' \emph{Nature},
  vol. 518, no. 7540, p. 529, 2015.

\bibitem{alpha-go-zero}
D.~Silver, J.~Schrittwieser, K.~Simonyan, I.~Antonoglou, A.~Huang, A.~Guez,
  T.~Hubert, L.~Baker, M.~Lai, A.~Bolton \emph{et~al.}, ``{Mastering the game
  of Go without human knowledge},'' \emph{Nature}, vol. 550, no. 7676, p. 354,
  2017.

\bibitem{alpha-zero}
D.~Silver, T.~Hubert, J.~Schrittwieser, I.~Antonoglou, M.~Lai, A.~Guez,
  M.~Lanctot, L.~Sifre, D.~Kumaran, T.~Graepel \emph{et~al.}, ``{Mastering
  Chess and Shogi by self-play with a general reinforcement learning
  algorithm},'' \emph{arXiv preprint arXiv:1712.01815}, 2017.

\bibitem{PPO}
J.~Schulman, F.~Wolski, P.~Dhariwal, A.~Radford, and O.~Klimov, ``Proximal
  policy optimization algorithms,'' \emph{arXiv preprint arXiv:1707.06347},
  2017.

\bibitem{openai-dota2}
{OpenAI Five}, 2018, [Online, June 2018]~\url{https://openai.com/five}.

\bibitem{cuccu2019playing}
G.~Cuccu, J.~Togelius, and P.~Cudr{\'e}-Mauroux, ``Playing atari with six
  neurons,'' in \emph{Proceedings of the 18th International Conference on
  Autonomous Agents and MultiAgent Systems}.\hskip 1em plus 0.5em minus
  0.4em\relax International Foundation for Autonomous Agents and Multiagent
  Systems, 2019, pp. 998--1006.

\bibitem{openai-gym}
{OpenAI Gym}, 2016, [Online]~\url{https://gym.openai.com}.

\bibitem{starcraft2}
O.~Vinyals, T.~Ewalds, S.~Bartunov, P.~Georgiev, A.~S. Vezhnevets, M.~Yeo,
  A.~Makhzani, H.~K{\"u}ttler, J.~Agapiou, J.~Schrittwieser \emph{et~al.},
  ``{StarCraft II: A new challenge for reinforcement learning},'' \emph{arXiv
  preprint arXiv:1708.04782}, 2017.

\bibitem{dopamine}
M.~G. Bellemare, P.~S. Castro, C.~Gelada, and S.~Kumar, [Online,
  2018]~\url{https://github.com/google/dopamine}.

\bibitem{rainbow}
M.~Hessel, J.~Modayil, H.~Van~Hasselt, T.~Schaul, G.~Ostrovski, W.~Dabney,
  D.~Horgan, B.~Piot, M.~Azar, and D.~Silver, ``Rainbow: Combining improvements
  in deep reinforcement learning,'' \emph{arXiv preprint arXiv:1710.02298},
  2017.

\bibitem{schulman2017proximal}
J.~Schulman, F.~Wolski, P.~Dhariwal, A.~Radford, and O.~Klimov, ``Proximal
  policy optimization algorithms,'' \emph{arXiv preprint arXiv:1707.06347},
  2017.

\bibitem{openai-es}
T.~Salimans, J.~Ho, X.~Chen, S.~Sidor, and I.~Sutskever, ``Evolution strategies
  as a scalable alternative to reinforcement learning,'' \emph{arXiv preprint
  arXiv:1703.03864}, 2017.

\bibitem{poupart-pomdp-continuous}
J.~Hoey and P.~Poupart, ``Solving {POMDPs} with continuous or large discrete
  observation spaces,'' in \emph{IJCAI}, 2005, pp. 1332--1338.

\bibitem{pomdp-continuous}
M.~T. Spaan and N.~Vlassis, ``{Perseus: Randomized point-based value iteration
  for POMDPs},'' \emph{Journal of artificial intelligence research}, vol.~24,
  pp. 195--220, 2005.

\bibitem{poupart-pomdp-continuous2}
J.~M. Porta, N.~Vlassis, M.~T. Spaan, and P.~Poupart, ``{Point-based value
  iteration for continuous POMDPs},'' \emph{Journal of Machine Learning
  Research}, vol.~7, no. Nov, pp. 2329--2367, 2006.

\bibitem{silver-cont-action}
T.~P. Lillicrap, J.~J. Hunt, A.~Pritzel, N.~Heess, T.~Erez, Y.~Tassa,
  D.~Silver, and D.~Wierstra, ``Continuous control with deep reinforcement
  learning,'' 2016.

\bibitem{stone-parameterized}
M.~Hausknecht and P.~Stone, ``Deep reinforcement learning in parameterized
  action space,'' \emph{arXiv preprint arXiv:1511.04143}, 2015.

\bibitem{prioritized-experience-replay}
T.~Schaul, J.~Quan, I.~Antonoglou, and D.~Silver, ``Prioritized experience
  replay,'' \emph{arXiv preprint arXiv:1511.05952}, 2015.

\bibitem{distributional-RL}
M.~G. Bellemare, W.~Dabney, and R.~Munos, ``A distributional perspective on
  reinforcement learning,'' in \emph{Proceedings of the 34th International
  Conference on Machine Learning-Volume 70}.\hskip 1em plus 0.5em minus
  0.4em\relax JMLR. org, 2017, pp. 449--458.

\bibitem{kemmerling2013making}
M.~Kemmerling, N.~Ackermann, and M.~Preuss, ``Making diplomacy bots
  individual,'' in \emph{Believable Bots}.\hskip 1em plus 0.5em minus
  0.4em\relax Springer, 2013, pp. 265--288.

\bibitem{GAIL}
J.~Ho and S.~Ermon, ``Generative adversarial imitation learning,'' in
  \emph{Advances in neural information processing systems}, 2016, pp.
  4565--4573.

\bibitem{IL1}
D.~A. Pomerleau, ``Alvinn: An autonomous land vehicle in a neural network,'' in
  \emph{Advances in neural information processing systems}, 1989, pp. 305--313.

\bibitem{IL2}
B.~D. Argall, S.~Chernova, M.~Veloso, and B.~Browning, ``A survey of robot
  learning from demonstration,'' \emph{Robotics and autonomous systems},
  vol.~57, no.~5, pp. 469--483, 2009.

\bibitem{IL3}
A.~Billard, S.~Calinon, R.~Dillmann, and S.~Schaal, ``Robot programming by
  demonstration,'' in \emph{Springer handbook of robotics}.\hskip 1em plus
  0.5em minus 0.4em\relax Springer, 2008, pp. 1371--1394.

\bibitem{thurau2007bayesian}
C.~Thurau, T.~Paczian, G.~Sagerer, and C.~Bauckhage, ``Bayesian imitation
  learning in game characters,'' \emph{International journal of intelligent
  systems technologies and applications}, vol.~2, no.~2, p. 284, 2007.

\bibitem{oneshot-IL}
Y.~Duan, M.~Andrychowicz, B.~Stadie, O.~J. Ho, J.~Schneider, I.~Sutskever,
  P.~Abbeel, and W.~Zaremba, ``One-shot imitation learning,'' in \emph{Advances
  in neural information processing systems}, 2017, pp. 1087--1098.

\bibitem{apprenticeship-learning}
P.~Abbeel and A.~Y. Ng, ``Apprenticeship learning via inverse reinforcement
  learning,'' in \emph{Proceedings of the twenty-first international conference
  on Machine learning}.\hskip 1em plus 0.5em minus 0.4em\relax ACM, 2004, p.~1.

\bibitem{IRL}
A.~Y. Ng, S.~J. Russell \emph{et~al.}, ``Algorithms for inverse reinforcement
  learning.'' in \emph{Icml}, 2000, pp. 663--670.

\bibitem{KaminskiMP}
\BIBentryALTinterwordspacing
M.~P. Kami{\'{n}}ski, ``In search of lexical discriminators of definition
  style: Comparing dictionaries through $n$-{G}rams,'' \emph{International
  Journal of Lexicography}, vol.~29, no.~4, pp. 403--423, 2016. [Online].
  Available: \url{http://dx.doi.org/10.1093/ijl/ecv038}
\BIBentrySTDinterwordspacing

\bibitem{davidwrite}
\BIBentryALTinterwordspacing
D.~Wright, ``Using word $n$-grams to identify authors and idiolects,''
  \emph{International Journal of Corpus Linguistics}, vol.~22, no.~2, pp.
  212--241, 2017. [Online]. Available:
  \url{http://www.jbe-platform.com/content/journals/10.1075/ijcl.22.2.03wri}
\BIBentrySTDinterwordspacing

\bibitem{andresen2017approximating}
M.~Andresen and H.~Zinsmeister, ``{Approximating Style by $N$-gram-based
  Annotation},'' in \emph{Proceedings of the Workshop on Stylistic Variation},
  2017, pp. 105--115.

\bibitem{RL_state_of_the_art}
M.~Wiering and van Martijn~Otterlo, \emph{Reinforcement Learning},
  1st~ed.\hskip 1em plus 0.5em minus 0.4em\relax Cambridge, MA, USA:
  Springer-Verlag Berlin Heidelberg, 2012, vol.~12.

\bibitem{digital_signal_proc}
A.~V. Oppenheim and R.~W. Schafer, \emph{Digital Signal Processing},
  1st~ed.\hskip 1em plus 0.5em minus 0.4em\relax Pearson, 1975.

\bibitem{vect_quant}
A.~Gersho and R.~M. Gray, \emph{Vector Quantization and Signal Compression},
  ser. Technology and Engineering.\hskip 1em plus 0.5em minus 0.4em\relax
  Springer Science and Business Media, 1991.

\bibitem{Pages}
G.~Pag{\`e}s, H.~Pham, and J.~Printems, ``Optimal quantization methods and
  applications to numerical problems in finance,'' in \emph{Handbook of
  computational and numerical methods in finance}.\hskip 1em plus 0.5em minus
  0.4em\relax Springer, 2004, pp. 253--297.

\bibitem{comp_graph}
J.~F. Hughes, A.~V. Dam, M.~McGuire, D.~F. Sklar, J.~D. Foley, S.~K. Feiner,
  and K.~Akeley, \emph{Computer Graphics}, 3rd~ed.\hskip 1em plus 0.5em minus
  0.4em\relax Addison-Wesley Professional, 2013.

\bibitem{dagger}
S.~Ross, G.~Gordon, and D.~Bagnell, ``A reduction of imitation learning and
  structured prediction to no-regret online learning,'' in \emph{Proceedings of
  the fourteenth international conference on artificial intelligence and
  statistics}, 2011, pp. 627--635.

\bibitem{async_rl}
V.~Mnih~et al., ``{Asynchronous Methods for Deep Reinforcement Learning},''
  \emph{arXiv:1602.01783v2}, 2016.

\end{thebibliography}
\end{document}